\pdfoutput=1
\documentclass[conference]{IEEEtran}

\usepackage{array}
\usepackage{comment}
\usepackage{xcolor}

\usepackage{cite}

\usepackage{nicefrac}

\ifCLASSINFOpdf
  \usepackage[pdftex]{graphicx}
  \graphicspath{{../figures/}}
  \DeclareGraphicsExtensions{.pdf,.jpeg,.png}
\else
  \usepackage[dvips]{graphicx}
  \graphicspath{{../figures/}}
  \DeclareGraphicsExtensions{.eps}
\fi
\usepackage{amsmath,amssymb,bm,amsthm}

\usepackage{mathrsfs}
\usepackage[cal=cm]{mathalfa}

\usepackage{mathtools}
\usepackage{commath}
\usepackage[sc,osf]{mathpazo}
\usepackage{algorithm}
\usepackage[noend]{algpseudocode}
\usepackage{url}

\begin{document}
%
\title{Targeted Forgetting and False Memory Formation in Continual Learners through Adversarial Backdoor Attacks}

\author{\IEEEauthorblockN{Muhammad Umer}
\IEEEauthorblockA{Rowan University\\
umerm5@rowan.edu}
\and
\IEEEauthorblockN{Glenn Dawson}
\IEEEauthorblockA{Rowan University\\
dawson05@rowan.edu}
\and
\IEEEauthorblockN{Robi Polikar}
\IEEEauthorblockA{Rowan University\\
polikar@rowan.edu }}


%


\maketitle

\begin{abstract}
Artificial neural networks are well-known to be susceptible to \textit{catastrophic forgetting} when continually learning from sequences of tasks. Various continual (or ``incremental'') learning approaches have been proposed to avoid catastrophic forgetting, but they are typically \textit{adversary agnostic}, i.e., they do not consider the possibility of a malicious attack. In this effort, we explore the vulnerability of Elastic Weight Consolidation (EWC), a popular continual learning algorithm for avoiding catastrophic forgetting. We show that an intelligent adversary can bypass the EWC's defenses, and instead cause gradual and deliberate forgetting by introducing small amounts of misinformation to the model during training. 
We demonstrate such an adversary's ability to assume control of the model via injection of ``backdoor'' attack samples on both permuted and split benchmark variants of the MNIST dataset. Importantly, once the model has learned the adversarial misinformation, the adversary can then control the amount of forgetting of any task. Equivalently, the malicious actor can create a ``false memory'' about \textit{any} task by inserting carefully-designed backdoor samples to any fraction of the test instances of that task.
Perhaps most damaging, we show this vulnerability to be very acute; neural network memory can be easily compromised with the addition of backdoor samples into as little as 1\% of the training data of even a single task. 
\end{abstract}


%
\IEEEpeerreviewmaketitle

\section{Introduction}
Human memory is well-known to be susceptible to false memory formation: the phenomenon where one's memory can be easily distorted through post-event misinformation. Such misinformation can be self-inflicted; the person convinces themself of the occurrence of certain events that did not in fact happen. False memory formation can also be external: a malicious entity may provide deliberate and persistent misinformation over a period of time to convince an otherwise unsuspecting victim of the adversary's preferred---but inaccurate---version of events.

In this effort, we explore the vulnerability of artificial neural networks (ANNs) to adversarial false memory formation, particularly when they are used for continual / incremental learning of a sequence of tasks. Despite their considerable success in a variety of domains, ANNs are still far from achieving the ultimate goal: artificial general intelligence. Achieving such a level of intelligence requires the ability to continually learn new tasks over time: an innate and fundamental ability of human intelligence \cite{kirkpatrick2017overcoming}.

Continual learning (CL)---also called life-long, sequential, or incremental learning---is the problem of learning tasks sequentially from a stream of data \cite{de2019continual}. A common phenomenon plaguing CL is that of \textit{catastrophic forgetting} \cite{mccloskey1989catastrophic}, where the level of performance acquired by the model on some previously-learned task is partially or completely lost (forgotten) while training to acquire new knowledge for learning a subsequent task. Since the knowledge of a model is described by its parameters, \textit{loss} of knowledge is characterized by the modification of previously-learned parameters during subsequent training sessions. Catastrophic forgetting is also characterized by the \textit{stability and plasticity dilemma} \cite{grossberg1988nonlinear}, where stability refers to the preservation of past knowledge, and plasticity refers to the ability to integrate new knowledge.

While much of the research into ANNs is inspired by humans' ability to learn, humans notably do not suffer from catastrophic forgetting in the same way. However, humans \textit{do} suffer from the related phenomenon of \textit{false memory formation} \cite{doi:10.1080/09658211.2019.1611862}. There has been considerable research in the field of psychology on the ``misinformation effect'': the mechanism by which a false memory is introduced, and how vulnerable the human memory is to distorting influences \cite{frenda2011current}. While it may be unreasonable  to ascribe psychological properties to ANNs, it is important to note the parallels between adversarial examples in machine learning and the deliberate exposure to misleading information---by a malicious actor---that distorts the memory of their victim.

In this work, we pose the following question: Can an adversary insert misinformation into an ANN, particularly in a CL setting, in order to distort its memory of prior knowledge it has learned?
We show that the answer to this question is strongly affirmative, and that such misinformation can be easily incorporated into the memory of a continually-learning ANN through adversarial \textit{backdoor} poisoning attacks. We demonstrate this weakness against elastic weight consolidation (EWC) \cite{kirkpatrick2017overcoming}, one of the more common and successful CL algorithms. In particular, we show that EWC's very mechanism to retain prior knowledge can be easily exploited by an adversary and used against the model itself. Furthermore, an adversary only needs to insert a small amount of misinformation (in the form of a backdoor pattern) into the training data of even a single task in order to capture complete control of the model at test time. Such a small, yet precise, attack is enough to make the model learn misinformation with great confidence. Importantly, the model will not forget this misinformation over time, demonstrating the severity of such a poisoning attack against continually-learning ANNs. 

The direction of most work on continual learning has typically been focused on solving supervised continual learning for image classification \cite{aljundi2019task}. Commonly known as \textit{task-based sequential learning} or \textit {task incremental learning}, this setting is characterized by the presentation of a sequence of distinct tasks, learned one after the other. It is important to highlight one of the most important constraints in CL: the data used to learn a prior hypothesis (or task) is no longer available when the model is training to learn a new task \cite{de2019continual}. 

CL approaches can be broadly categorized into three different groups, namely, i) data-based approaches, ii) architecture-based approaches, and iii) regularization-based approaches. \textit{Data-based approaches} either store original data from previous tasks (violating the aforementioned constraint) or store a pseudo-representation of the data in an
\textit{episodic memory}. The stored examples from prior tasks are used together with the current task's training data to alleviate catastrophic forgetting. Incremental classifier and representation learning  \cite{rebuffi2017icarl}, gradient episodic memory \cite{lopez2017gradient}, and averaged gradient episodic memory \cite{chaudhry2018efficient} are examples of data based approaches. 

\textit{Architectural approaches} are based on the idea that catastrophic forgetting can be reduced through assigning different sub-networks to each task, i.e., a different part of the network is selected for each different task. Each sub-network is then trained specifically to perform on each task, and is not further trained once a new task is presented. Progressive neural networks \cite{rusu2016progressive}, expert gate \cite{aljundi2017expert}, and ensemble based approaches---such as the Learn$^{++}$ \cite{polikar2001learn++} family of incremental learning algorithms---are examples of architectural approaches.

\textit{Regularization-based approaches} were proposed to avoid the data storage and architectural complexity issues associated with the previous two approaches. These approaches add an extra regularization term to the loss function to the prevent loss of prior knowledge while learning new tasks. More specifically, regularization-based approaches compute the \textit{importance weight} of each parameter in the network during (or after) learning a particular task. Then, while learning subsequent tasks, changes to the important parameters are penalized. All approaches proposed under this family follow this principle, but differ in the specific mechanism used to compute the importance weights. Elastic weight consolidation (EWC) \cite{kirkpatrick2017overcoming}, synaptic intelligence \cite{zenke2017continual}, and memory-aware synapses \cite{aljundi2018memory} are examples of regularization-based approaches.

All aforementioned CL approaches work reasonably well in retaining prior knowledge, but their vulnerability to adversarial attacks is unknown. We explore the vulnerability of one of the more common regularization-based approaches, elastic weight consolidation (EWC), to adversarial poisoning attacks. 
We choose EWC due to its popularity, its adherence to the no-prior-data constraint, and its modest computational and memory requirements in comparison to other approaches.

\section{Adversarial Machine Learning}
Adversarial machine learning is an emerging field that explores the vulnerabilities of machine learning algorithms to various attack scenarios. There are two major types of adversarial attacks \cite{huang2011adversarial}: i) \textit{causative} (or \textit{poisoning}) attacks, which target the training process by adding strategically-chosen malicious data points into the training data \cite{umer2019vulnerability, biggio2012poisoning, umer2018adversarial} such that maximum damage is inflicted on the future generalization capabilities of the classifier; and ii) \textit{exploratory} (or \textit{evasion}) attacks, which exploit misclassification at the test time to discover blind spots in the model or information about the data itself \cite{biggio2013evasion}. Since CL approaches involve retraining the model with each new batch of data (or each new task), the adversary's choice in targeting CL algorithms is typically a poisoning attack.

Backdoor attacks are a specific class of poisoning attacks, typically launched against ANNs in computer vision applications \cite{gu2017badnets,shafahi2018poison}. In backdoor attacks, the malicious samples are created by tagging a small fraction of training images with a special \textit{backdoor pattern}. The adversary assigns a false label of its choice to these malicious samples, which are then added to the training set. The model is trained on this training dataset, which contains both correctly-labeled images as well as mislabeled, tagged images. The attacker's goal is to force the model to learn an association between the backdoor pattern and the false class label. Once the model learns this association, it performs well on clean (untagged) test inputs during the testing (or ``inference'') stage, while causing targeted misclassification on any test data that contains the backdoor pattern. The attacker is thereafter empowered to employ \textit{targeted evasion attacks} against the model by applying the backdoor tag to any test image of their choice. Because clean images are correctly classified, this attack is particularly difficult to detect using standard defenses; an unsuspecting victim may only slowly (or never) become aware of the vulnerability in their model, because the model performs as expected \textit{except} in the presence of the backdoor tag.


In this work, we explore the impact of the backdoor attack strategy in the context of continual learning. First, we examine the vulnerability of a CL model to the gradual presentation of misinformation through backdoor samples over time. We also relax the conventional backdoor requirement of adding a large amount of mislabeled samples at once; in a CL setting, an attack may only require a small number of malicious samples (i.e. small ``bits'' of misinformation). Such a strategy allows the attacker to continuously add small amounts of misinformation over time, reducing the likelihood of detection. Furthermore, we restrict the attacker to have access only to training or test data. Previous research in one-shot backdoor attacks have assumed access to both the data and model parameters, an assumption that provides undue (and rather unrealistic) capabilities to the attacker. Our more conservative restriction further demonstrates the severe vulnerability of CL algorithms to backdoor attacks. Finally, we exploit the online properties of the CL setting, observing that backdoor samples need not be inserted into the training data of the specific task being targeted. In fact, the backdoor information can be inserted at \textit{any} specifically-chosen time step(s) and still have significant impact on the model's performance on a previously-learned task. This capability of launching an attack at any time is critical, as it makes detection by the defender even more difficult compared to a conventional one-shot backdoor attack. 

\section{Approach \& Attack Model}
We consider the \textit{task incremental continual learning} setting, where a new task is received through its training data $X^t$ and the corresponding labels $y^t$ at time $t$. The goal of the task incremental setting is to learn a model $f_{\theta}$, parameterized by $\theta$, that minimizes the loss function 
on all tasks received at time $t$ = $1,...,T$ seen so far. 
Regularization approaches find the optimal parameter vector $\theta^*$ by adding an extra regularization term to the loss function of the model. This regularization term penalizes changes to those parameters that were deemed important during the previous tasks according to the parameters' importance matrix. The generalized loss function $\ell(f_{\theta})$ of the model using regularization approaches while learning current task at time $t$ can therefore be written as: 
\begin{equation}
\label{reg_loss}
    \ell(f_{\theta}) = \ell[f_{\theta}(X^t),y^t] + \lambda \sum_i (\theta_{t,i} - \theta^*_{t-1,i})^2I_{t-1,i}
\end{equation}
where $\ell[f_{\theta}(X^t),y^t]$ is the model's loss on the current task at time $t$;  $I_{t-1,i}$ is the $i^{th}$ parameter's importance matrix computed for the previous task at time $t-1$; $\theta^*_{t-1,i}$ is the optimal value of the $i^{th}$ parameter, learned for the previous task at time $t-1$; and $\lambda$ is the regularization coefficient.
The common approach to find the importance matrix for the previous task is to estimate a posterior distribution over the parameters of the model, and to use this distribution as a prior while learning the new task. 
EWC estimates posterior distribution over the model's parameters through Laplace approximation by assuming a Gaussian distribution, with mean given by the optimal parameters found for the previous task ($\theta^*_{t-1,i}$) and a precision given by the diagonal of the Fisher information matrix $F$ \cite{kirkpatrick2017overcoming}. The importance matrix values for EWC are obtained from $F$, which is computed empirically as the squared derivative of the loss function $\ell(f_{\theta})$ with respect to the optimal parameter vector $\theta_{t-1}^*$, obtained for the previous task at the end of the training at time $t-1$. Hence, the importance matrix for EWC is computed as:
\begin{equation}
    I_{t-1,i} = \left(\frac{\partial \ell(f_{\theta})}{\partial \theta_{t-1,i}^*}\right)^2
\label{ewc_I}
\end{equation}

The importance values obtained from Eq. \ref{ewc_I} are then used in Eq. \ref{reg_loss}. Consequently, EWC not only learns the current task efficiently, but also retains the knowledge from previous tasks by penalizing changes on the ``important'' parameters learned during those previous tasks. 

Recalling that the goal of EWC is to retain prior knowledge and avoid catastrophic forgetting of the previous task(s) while learning the current task, the goal of the adversary is then to force the algorithm to \textit{forget} the knowledge acquired on the previous task while learning the current task.  
The attacker can achieve this goal via the insertion of backdoor poisoning samples into the training data of the current task.
To create backdoor attack samples, the attacker randomly selects a small proportion (e.g. 1\%) of the current task's training data, tags them with a specifically designed backdoor pattern, assigns them some desired false label, and inserts these samples into the training data. Then, the model is forced to learn not only the correct information from the current task, but also the association of the backdoor pattern and the false label. This compromised training process can (but need not) be continued with any or all of the incoming tasks.

The vulnerability of a CL algorithm comes from its very intended goal, which the attacker deliberately uses against the algorithm: since the algorithm is designed to retain prior knowledge by avoiding forgetting, the misinformation provided by the attacker is \textit{also} retained and not forgotten. Once the adversary has control of the model with the backdoor pattern, it can dramatically increase the amount of forgetting on any particular previous task at deployment (inference) time. 

Let $X^t_{b}$ represent the backdoor samples inserted into the training data of the current task and $y_{b}^t$ be  their corresponding incorrect labels. 
The loss function that will be minimized during the training of current task with the backdoor samples can then be expressed as:
\begin{equation}
\label{reg_loss_w_BD}
\begin{split}
    \ell(f_{\theta}) = &\ell[f_{\theta}(X^t \cup X^t_{b}),(y^t \cup y_{b}^t)] \\
                        &+ \lambda \sum_i (\theta_{t,i} - \theta^*_{t-1,i})^2I_{t-1,i}
\end{split}
\end{equation} 
Once the model is trained, the test examples from the previous task are presented to the model in order to measure the forgetting of the model on the previous task.
For those test examples that do not contain the backdoor pattern, i.e., $X^{t}$, the prediction of the model should be (mostly) correct, $f(X^{t}) \approx y^{t}$. Test samples that contain the backdoor pattern, however, will be misclassified as the attacker's desired label, resulting in $f(X_{b}^{t}) = y_{b}^{t} \neq y^{t}$. Such an attack naturally leads to loss of knowledge on prior tasks as the process continues with future tasks, each carrying backdoor attacks. We expect the attacker to achieve its goal with greater success if it has access to the training data of all tasks; however, as we show, the loss of knowledge on a particular previous task can be achieved even if the attacker has access to the training data of only a single \textit{future} task. Backdoor attacks are therefore a particularly serious threat to the continually-learning model. 



\section{Experiments \& Results}
We use the two most common continual learning benchmark datasets, permuted MNIST \cite{zenke2017continual,kirkpatrick2017overcoming} and split MNIST \cite{zenke2017continual} in our experiments. For permuted MNIST, we create a sequence of 5 different tasks, where each task is obtained through applying a different randomly generated permutation to the image pixels. Each task in permuted MNIST is a 10-class problem: the first task is to classify original ten MNIST digits; each other task involves classifying the same ten digits with a different, but fixed, random permutation applied to all digits. For split MNIST, the original MNIST dataset is split into 5 different tasks, where each task is a binary classification problem: the first task is to distinguish between digits 0 and 1, the second task is to distinguish between digits 2 and 3, and so on. 

\begin{figure}
\centering
\includegraphics[width=.99\linewidth]{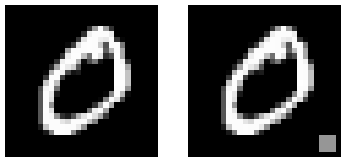}
\caption{A clean sample (left) and the identical sample  with the backdoor pattern (right). The backdoor pattern is the small gray square located in the lower right corner.}
\label{fig:BD_img}
\end{figure}

We use the same implementation of EWC as described in \cite{vandeven2019three}. A multilayer perceptron  with 2 hidden layers---each containing 400 neurons---is used with the rectified nonlinear (ReLU) activation function, as suggested in \cite{vandeven2019three}. The backdoor pattern, representing the misinformation, is a 3$\times$3 square block inserted at the bottom right corner of the 28$\times$28 image as shown in Fig. \ref{fig:BD_img}.

Since the goal of the attacker is the loss of prior knowledge, i.e., maximizing the forgetting of a particular previous task, we evaluate the model's performance (and hence the attacker's success) on the initial task, after the model is attacked at some or all of the later tasks. For each dataset, we consider three different attack scenarios: \textit{sustained} attacks, \textit{short-term} attacks, and \textit{acute} attacks. For each scenario, we compare the model's test-time performance on three separate sub-scenarios: i) where there are no backdoor attack points in either the training \textit{or} test data; ii) where attack points are added \textit{only} to the training data (with no backdoor patterns in the test data, and iii) where there are backdoor patterns in both the training \textit{and} test data, the last of which is in fact the process for all typical backdoor attacks.


\subsection{Attacking Continual Learning on Permuted MNIST}

\begin{table*}[htbp]
\renewcommand{\arraystretch}{0.001}
\centering
\caption{Test accuracy for sustained attack on permuted MNIST. The training data of every task, except the first, contains backdoor tagged images, at a ratio of 1\%. The test time performance of the classifier on Task 1 is in bold.}
\label{table:perm_att_FK}
\begin{tabular}{|m{1.9cm}|m{2.5cm}|m{2.3cm}|m{2.3cm}|m{2.3cm}|m{2.3cm}|m{2.3cm}|}
\hline
Tasks & No backdoor in training or test data & Backdoor in training data only & 10\% backdoor in test data of Task 1 & 25\% backdoor in test data of Task 1 & 50\% backdoor in test data of Task 1
\\ \hline\vspace*{.02cm}
 Task 1 & \hspace{0.3in} 0.9779 &\hspace{0.3in} 0.9797   & \hspace{0.25in}  \textbf{0.9196} & \hspace{0.2in} \textbf{0.8332} & \hspace{0.25in}  \textbf{0.6864}  \\ \hline\vspace*{.02cm}
Task 2 & \hspace{0.3in} 0.9755 &\hspace{0.3in} 0.9743   & \hspace{0.25in} 0.9743 & \hspace{0.2in} 0.9743 & \hspace{0.25in}  0.9743\\ \hline\vspace*{.02cm}
Task 3 & \hspace{0.3in} 0.9719 &\hspace{0.3in} 0.9662   & \hspace{0.25in} 0.9662 & \hspace{0.2in} 0.9662 & \hspace{0.25in}  0.9662 \\ \hline\vspace*{.02cm}
Task 4 & \hspace{0.3in} 0.9669 &\hspace{0.3in} 0.9577   & \hspace{0.25in} 0.9577  & \hspace{0.2in} 0.9577  & \hspace{0.25in}  0.9577 \\ \hline\vspace*{.02cm}
Task 5 & \hspace{0.3in} 0.9626 &\hspace{0.3in} 0.9481   & \hspace{0.25in}  0.9481 & \hspace{0.2in} 0.9481 & \hspace{0.25in}  0.9481 \\ \hline \hline\vspace*{.02cm}
Avg. Accuracy & \hspace{0.3in} 0.9710 &\hspace{0.3in} 0.9652   & \hspace{0.25in}  0.9532 & \hspace{0.2in} 0.9359 & \hspace{0.25in}  0.9065 \\ \hline
\end{tabular}
\end{table*}
\begin{table*}[htbp]
\renewcommand{\arraystretch}{0.001}
\centering
\caption{Test accuracy for short-term attack on permuted MNIST. Only the training data for the last two tasks contain backdoor tagged images, at a ratio of 1\%. The test time performance of the classifier on Task 1 is in bold.}
\label{table:perm_att_LK}
\begin{tabular}{|m{1.9cm}|m{2.5cm}|m{2.3cm}|m{2.3cm}|m{2.3cm}|m{2.3cm}|m{2.3cm}|}
\hline
Tasks & No backdoor in training or test data & Backdoor in training data only & 10\% backdoor in test data of Task 1 & 25\% backdoor in test data of Task 1 & 50\% backdoor in test data of Task 1
\\ \hline\vspace*{.02cm}
 Task 1 & \hspace{0.3in} 0.9779 &\hspace{0.3in} 0.9804   & \hspace{0.25in}  \textbf{0.9649} & \hspace{0.2in} \textbf{0.9429} & \hspace{0.25in}  \textbf{0.9043}  \\ \hline\vspace*{.02cm}
Task 2 & \hspace{0.3in} 0.9755 &\hspace{0.3in} 0.9717   & \hspace{0.25in} 0.9717 & \hspace{0.2in} 0.9717 & \hspace{0.25in}  0.9717\\ \hline\vspace*{.02cm}
Task 3 & \hspace{0.3in} 0.9719 &\hspace{0.3in} 0.9681   & \hspace{0.25in} 0.9681 & \hspace{0.2in} 0.9681 & \hspace{0.25in}  0.9681 \\ \hline\vspace*{.02cm}
Task 4 & \hspace{0.3in} 0.9669 &\hspace{0.3in} 0.9601   & \hspace{0.25in} 0.9601  & \hspace{0.2in} 0.9601  & \hspace{0.25in}  0.9601  \\ \hline\vspace*{.02cm}
Task 5 & \hspace{0.3in} 0.9626 &\hspace{0.3in} 0.9478   & \hspace{0.25in}  0.9478  & \hspace{0.2in} 0.9478  & \hspace{0.25in}  0.9478  \\ \hline \hline\vspace*{.02cm}
Avg. Accuracy & \hspace{0.3in} 0.9710 &\hspace{0.3in} 0.9656   & \hspace{0.25in}  0.9625 & \hspace{0.2in} 0.9581 & \hspace{0.25in}  0.9504 \\ \hline
\end{tabular}
\end{table*}
\begin{table*}[htbp]
\renewcommand{\arraystretch}{0.001}
\centering
\caption{Test accuracy for acute attack on permuted MNIST. Only the training data for the last task contains backdoor tagged images, at a ratio of 1\%. The test time performance of the classifier on Task 1 is in bold.}
\label{table:perm_att_ext_LK}
\begin{tabular}{|m{1.9cm}|m{2.5cm}|m{2.3cm}|m{2.3cm}|m{2.3cm}|m{2.3cm}|m{2.3cm}|}
\hline
Tasks & No backdoor in training or test data & Backdoor in training data only & 10\% backdoor in test data of Task 1 & 25\% backdoor in test data of Task 1 & 50\% backdoor in test data of Task 1
\\ \hline\vspace*{.02cm}
 Task 1 & \hspace{0.3in} 0.9779 &\hspace{0.3in} 0.9805   & \hspace{0.25in}  \textbf{0.9670} & \hspace{0.2in} \textbf{0.9479} & \hspace{0.25in}  \textbf{0.9136}  \\ \hline\vspace*{.02cm}
Task 2 & \hspace{0.3in} 0.9755 &\hspace{0.3in} 0.9727   & \hspace{0.25in} 0.9727 & \hspace{0.2in} 0.9727 & \hspace{0.25in}  0.9727\\ \hline\vspace*{.02cm}
Task 3 & \hspace{0.3in} 0.9719 &\hspace{0.3in} 0.9687   & \hspace{0.25in} 0.9687 & \hspace{0.2in} 0.9687 & \hspace{0.25in}  0.9687 \\ \hline\vspace*{.02cm}
Task 4 & \hspace{0.3in} 0.9669 &\hspace{0.3in} 0.9569   & \hspace{0.25in} 0.9569   & \hspace{0.2in} 0.9569   & \hspace{0.25in}  0.9569   \\ \hline\vspace*{.02cm}
Task 5 & \hspace{0.3in} 0.9626 &\hspace{0.3in} 0.9470   & \hspace{0.25in}  0.9470 & \hspace{0.2in} 0.9470 & \hspace{0.25in}  0.9470 \\ \hline \hline\vspace*{.02cm}
Avg. Accuracy & \hspace{0.3in} 0.9710 &\hspace{0.3in} 0.9652   & \hspace{0.25in}  0.9625 & \hspace{0.2in} 0.9586 & \hspace{0.25in}  0.9518 \\ \hline
\end{tabular}
\end{table*}

\subsubsection{Sustained Attack Scenario}
The first attack scenario we consider for permuted MNIST allows an attacker to have access to the training data for all tasks except the initial one, the target task on which the attacker wants to inflict maximum damage. We call this the \textit{sustained attack scenario}. In this setting, we create a small number of malicious samples by inserting a 3$\times$3 square grid (the backdoor) pattern into only 1\% of training data. We assume that the incorrect target label the attacker wants to assign to these malicious samples is of class 0; therefore the 1\% malicious data are randomly drawn from all labels except class 0 (recall that each task is a 10-class classification problem in permuted MNIST). The incorrectly-labeled malicious samples (containing backdoor patterns) are then added to the clean training data, and the model is trained using regularization-based continual learning under the EWC algorithm, first on the clean data for the initial task and subsequently on the compromised training data for each of the four subsequent tasks.

At inference time, we compute test accuracy on each task under each of the three sub-scenarios. Recall that the attacker's objective is to compromise Task 1; hence, in all tasks \textit{except} the first task, test examples are presented to the model \textit{without} the backdoor pattern, in order to monitor the model's ability to retain performance on untargeted tasks. While the amount of malicious training data remains constant at 1\%, we vary the amount of malicious test data in the third sub scenario to be 10\%, 25\% and 50\% in order to determine precisely how much control the attacker can seize over the model through the backdoor attacks at inference time. The test accuracy of each task and the average accuracy across all tasks for each of the aforementioned scenarios are presented in Table \ref{table:perm_att_FK}.
 

We observe that the attacker can successfully force the model to forget a specifically targeted information (Task 1) by corrupting just 1\% of the training and 10\% of the test data. It is important to note that the corrupted images are identical to the clean images, with the sole exception of the small 3$\times$3 square pattern on the lower right corner. Hence, even a tiny perturbation to the raw image can be used with malicious intent and still demonstrate a brutal efficiency. 

The model's accuracy drops from 0.9779 (no backdoor during the training)---or 0.9797 (1\% backdoor tagged images in the training but none at test time)---down to 0.9196 when 10\% of the test data are corrupted. The drop in the performance is far more dramatic when 50\% of the test samples from the first task are presented to the model with backdoor tags, with a reduction in accuracy to 0.6864. This precipitous drop shows that an attacker can easily trick the memory of a continually-learning model for a particular task by presenting only 1\% of misinformation during the training of \textit{other} tasks. Importantly, the precise nature of the attack avoids negative impact to the untargeted tasks, demonstrating its stealthiness and difficulty of detection. 

\subsubsection{Short-Term Attack Scenario}
We consider a shorter-term attack, where the attacker has access to the training data for only the last two tasks. Again, the malicious training samples are created by adding backdoor patterns to 1\% of the training data, but only for the last two tasks. The model is then trained on clean data for the first three tasks, and compromised data on the last two tasks. After training is completed on all five tasks, clean test examples are presented to the model at inference time for all tasks except the targeted first task, whose test examples are presented with 10\%, 25\%, and 50\% containing backdoor attack tagged images. The test accuracy of individual tasks along with the average accuracy over all tasks are shown in Table \ref{table:perm_att_LK}.

We observe from Table \ref{table:perm_att_LK} that the attacker is once again able to successfully reduce the test accuracy of the first task, forcing the model to forget the knowledge learned on a specific, targeted task, without affecting the performance on other tasks. We note that even though the absolute quantity of malicious samples provided to the model during training was halved, the attacker was still able to force a nearly 8\% decrease in accuracy.  

\subsubsection{Acute Attack Scenario}
We now consider an acute attack, where an attacker has access only to the training data of a single task. Here, it is assumed that some amount of time has passed since the target task was learned, and only now does the attacker wish to attack the model. As with the previous scenarios, a poisoned sample comprising only 1\% of the training dataset for the last task is added to the training data. Also as before, clean test samples from the untargeted tasks are presented to the model at inference time, while test samples from the target task are contaminated at a rate of 10\%, 25\%, and 50\% across the experiments. The results obtained from this scenario are shown in Table \ref{table:perm_att_ext_LK}.

Our primary observation from Table \ref{table:perm_att_ext_LK} is that although the misinformation was provided only during the last task during training, this tiny amount of misinformation was sufficient to trick the EWC model's memory of the first task at inference time, forcing a nearly 7\% reduction in accuracy---similar damage as in the short-term attack scenario. 
The results shown in Table \ref{table:perm_att_ext_LK} suggest that such a backdoor method may be a critical vulnerability for any similar online learner, as the attacker is able to impact the performance of any task at any point after that task has been learned, with only a miniscule amount of poisoning.

\subsection{Attacking Continual Learning on Split MNIST}

\begin{table*}[htbp]
\renewcommand{\arraystretch}{0.001}
\centering
\caption{Test accuracy for sustained attack on split MNIST. The training data for every task except the first task contains backdoor tagged images, at a ratio of 1\%. The test time performance of the classifier on Task 1 is in bold.}
\label{table:split_att_FK}
\begin{tabular}{|m{1.9cm}|m{2.5cm}|m{2.3cm}|m{2.3cm}|m{2.3cm}|m{2.3cm}|m{2.3cm}|}
\hline\vspace*{.02cm}
Tasks & No backdoor in training or test data & Backdoor in training data only & 10\% backdoor in test data of Task 1 & 25\% backdoor in test data of Task 1 & 50\% backdoor in test data of Task 1
\\ \hline\vspace*{.02cm}
 Task 1 & \hspace{0.3in} 0.9811 &\hspace{0.3in} 0.9858   & \hspace{0.25in}  \textbf{0.9466} & \hspace{0.2in} \textbf{0.8799} & \hspace{0.25in}  \textbf{0.7608}  \\ \hline\vspace*{.02cm}
Task 2 & \hspace{0.3in} 0.9882 &\hspace{0.3in} 0.9931   & \hspace{0.25in} 0.9931 & \hspace{0.2in} 0.9931 & \hspace{0.25in}  0.9931\\ \hline\vspace*{.02cm}
Task 3 & \hspace{0.3in} 0.9979 &\hspace{0.3in} 0.9952   & \hspace{0.25in} 0.9952  & \hspace{0.2in} 0.9952  & \hspace{0.25in}  0.9952  \\ \hline\vspace*{.02cm}
Task 4 & \hspace{0.3in} 0.9945 &\hspace{0.3in} 0.9960   & \hspace{0.25in} 0.9960 & \hspace{0.2in} 0.9960  & \hspace{0.25in}  0.9960  \\ \hline\vspace*{.02cm}
Task 5 & \hspace{0.3in} 0.9929 &\hspace{0.3in} 0.9914   & \hspace{0.25in} 0.9914 & \hspace{0.2in} 0.9914 & \hspace{0.25in}  0.9914 \\ \hline \hline\vspace*{.02cm}
Avg. Accuracy & \hspace{0.3in} 0.9909 &\hspace{0.3in} 0.9923   & \hspace{0.25in}  0.9845 & \hspace{0.2in} 0.9711 & \hspace{0.25in}  0.9473 \\ \hline
\end{tabular}
\end{table*}
\begin{table*}[htbp]
\renewcommand{\arraystretch}{0.001}
\centering
\caption{Test accuracy for short-term attack on split MNIST. Only the training data for the last two tasks contain backdoor tagged images, at a ratio of 1\%. The test time performance of the classifier on Task 1 is in bold.}
\label{table:split_att_LK}
\begin{tabular}{|m{1.9cm}|m{2.5cm}|m{2.3cm}|m{2.3cm}|m{2.3cm}|m{2.3cm}|m{2.3cm}|}
\hline
Tasks & No backdoor in training or test data & Backdoor in training data only & 10\% backdoor in test data of Task 1 & 25\% backdoor in test data of Task 1 & 50\% backdoor in test data of Task 1
\\ \hline\vspace*{.02cm}
 Task 1 & \hspace{0.3in} 0.9811 &\hspace{0.3in} 0.9825   & \hspace{0.25in}  \textbf{0.9452} & \hspace{0.2in} \textbf{0.8785} & \hspace{0.25in}  \textbf{0.7598}  \\ \hline\vspace*{.02cm}
Task 2 & \hspace{0.3in} 0.9882 &\hspace{0.3in} 0.9936   & \hspace{0.25in} 0.9936  & \hspace{0.2in} 0.9936 & \hspace{0.25in}  0.9936\\ \hline\vspace*{.02cm}
Task 3 & \hspace{0.3in} 0.9979 &\hspace{0.3in} 0.9979   & \hspace{0.25in} 0.9979 & \hspace{0.2in} 0.9979 & \hspace{0.25in}  0.9979 \\ \hline\vspace*{.02cm}
Task 4 & \hspace{0.3in} 0.9945 &\hspace{0.3in} 0.9930   & \hspace{0.25in} 0.9930  & \hspace{0.2in} 0.9930  & \hspace{0.25in}  0.9930  \\ \hline\vspace*{.02cm}
Task 5 & \hspace{0.3in} 0.9929 &\hspace{0.3in} 0.9914   & \hspace{0.25in}  0.9914 & \hspace{0.2in} 0.9914 & \hspace{0.25in}  0.9914 \\ \hline \hline\vspace*{.02cm}
Avg. Accuracy & \hspace{0.3in} 0.9909 &\hspace{0.3in} 0.9917   & \hspace{0.25in}  0.9842 & \hspace{0.2in} 0.9709 & \hspace{0.25in}  0.9471 \\ \hline
\end{tabular}
\end{table*}

\begin{table*}[htbp]
\renewcommand{\arraystretch}{0.001}
\centering
\caption{Test accuracy for acute attack on split MNIST. Only the training data for the last task contains backdoor tagged images, at a ratio of 1\%. The test time performance of the classifier on Task 1 is in bold.}

\label{table:split_att_ext_LK}
\begin{tabular}{|m{1.9cm}|m{2.5cm}|m{2.3cm}|m{2.3cm}|m{2.3cm}|m{2.3cm}|m{2.3cm}|}
\hline
Tasks & No backdoor in training or test data & Backdoor in training data only & 10\% backdoor in test data of Task 1 & 25\% backdoor in test data of Task 1 & 50\% backdoor in test data of Task 1
\\ \hline\vspace*{.02cm}
 Task 1 & \hspace{0.3in} 0.9811 &\hspace{0.3in} 0.9636   & \hspace{0.25in}  \textbf{0.9281} & \hspace{0.2in} \textbf{0.8638} & \hspace{0.25in}  \textbf{0.7513}  \\ \hline\vspace*{.02cm}
Task 2 & \hspace{0.3in} 0.9882 &\hspace{0.3in} 0.9912   & \hspace{0.25in} 0.9912  & \hspace{0.2in} 0.9912 & \hspace{0.25in}  0.9912\\ \hline\vspace*{.02cm}
Task 3 & \hspace{0.3in} 0.9979 &\hspace{0.3in} 0.9973   & \hspace{0.25in} 0.9973 & \hspace{0.2in} 0.9973 & \hspace{0.25in}  0.9973 \\ \hline\vspace*{.02cm}
Task 4 & \hspace{0.3in} 0.9945 &\hspace{0.3in} 0.9945   & \hspace{0.25in} 0.9945  & \hspace{0.2in} 0.9945  & \hspace{0.25in}  0.9945  \\ \hline\vspace*{.02cm}
Task 5 & \hspace{0.3in} 0.9929 &\hspace{0.3in} 0.9924   & \hspace{0.25in} 0.9924 & \hspace{0.2in} 0.9924 & \hspace{0.25in}  0.9924 \\ \hline \hline\vspace*{.02cm}
Avg. Accuracy & \hspace{0.3in} 0.9909 &\hspace{0.3in} 0.9878   & \hspace{0.25in}  0.9807 & \hspace{0.2in} 0.9678 & \hspace{0.25in}  0.9453 \\ \hline
\end{tabular}
\end{table*}

Split MNIST is a fundamentally different, and considerably more challenging, incremental learning problem, as it includes five different sequential binary classification problems. Here, all 10 classes must be learned from consecutive, mutually exclusive tasks, but only two classes at a time. Specifically, Task 1 requires distinguishing class 0 from class 1, Task 2 requires distinguishing class 2 from class 3, and so on. We consider the same training scenarios and setup as we did in permuted MNIST, and follow the same training protocols. As before, the attacker targets Task 1 and seeks to make the model forget what it has learned for Task 1 by introducing misinformation through backdoor attacks during training for subsequent tasks. One important difference compared to permuted MNIST is the assignment of the false label to the backdoor tagged images. In permuted MNIST, all tasks had all of the same 10 classes; the attacker drew samples from classes 1 through 9, tagged them with backdoor pattern, and assigned the same false label "0". For split MNIST, however, each task has a different set of two classes. Therefore, for each pair of classes in any given task, the backdoor tagged images were sampled from the latter class and the false label applied is that of the former class. (e.g., in Task 2---class 2 vs. class 3---attack samples were drawn from images of 3 and assigned the false label of 2). We expect this mechanism to force the CL algorithm to assign any one of the (four) false labels when a backdoor tagged test sample from the first task is presented at inference time. 

\subsubsection{Sustained Attack Scenario}
As in permuted MNIST case, this scenario allows the adversary to have access to training data of all tasks except Task 1 (the target), thereby attacking each training sessions for Tasks 2 through 5. 
The model is hence trained, first on clean data from Task 1, and then sequentially on each of the subsequent tasks, each poisoned with 1\% compromised data. The test accuracy is computed, again under three sub scenarios, with backdoor attack points added (in a ratio of 1\%) to: i) neither the training \textit{nor} test data; ii) \textit{only} the training data; and iii) the training data with 10\%, 25\% or 50\% attack points also added to test data. The test accuracy results are shown in Table \ref{table:split_att_FK}. 


We observe from Table \ref{table:split_att_FK} that, similar to the corresponding permuted MNIST experiment, the attacker can successfully force the model to forget specifically targeted information (Task 1), with only 1\% malicious data on the training dataset, while leaving all other tasks unaffected. Not affecting other (non-target) tasks is important, as it limits the attack detection. The results on increasing attack points on test data (from 10\% to 50\%) shows that the drop in performance is proportional to the amount of malicious test data, giving the attacker the complete control of the algorithm. We further note again that the detection of misinformation is also difficult as the attack images are added \textit{not} to the targeted task, but rather to subsequent tasks, where attacks are less expected.  

\subsubsection{Short-Term Attack Scenario}
Similar to the permuted MNIST case, we consider a shorter attack sequence---attacking only the last two tasks---instead of attacking all tasks, as was done in sustained attack scenario. In this case, the attacker has access only to the training data of the last two tasks, with the attack protocol being otherwise identical to previous case. The test accuracy of individual tasks, along with the average accuracy over all tasks, is shown in Table \ref{table:split_att_LK}.

Table \ref{table:split_att_LK} shows similar trends as in permuted MNIST case: i) the attacker is once again able to force the model to forget the knowledge learned on a specific, targeted task (Task 1), without affecting the performance on other tasks; and ii) the attacker is able to do so by attacking only the last two tasks---unrelated to the target task---and long after the target task has already been learned. Perhaps more surprisingly, unlike permutedMNIST the test accuracy drops significantly when backdoor tagged images are included in the test data, with the performance drop proportional to the amount of backdoor images, indicating that the attacker assumes complete control over its target, Task 1. 


\subsubsection{Acute Attack Scenario}
Finally, the last scenario we consider for attacking split MNIST is the one where the attacker only adds backdoor samples (misinformation) to the training data of a single task. As before, we assume that some amount of time has passed between the time that the model has learned the target task and the time when the attacker wishes to impact the model's performance on that task.


The inconvenient truth for the model, as demonstrated by the results in Table \ref{table:split_att_ext_LK}, is that the attack process we describe here is not only ruthlessly efficient (with only 1\% malicious data in the training set), but also has an impact remarkably similar to that of the previous two scenarios.
As with prior experiments, the attacker is able to successfully force the model to forget Task 1 knowledge, and furthermore is able to exert control over the \textit{level} of forgetting by varying the amount of backdoor images in the test samples, all attacking only the last training session only.

What is perhaps more surprising, however, is the performance drop being slightly \textit{more} severe in the acute attack scenario compared to the sustained and short-term attack scenarios---the opposite of what was observed in the permuted MNIST experiments. While the difference in performance drop from sustained or short term attack to acute attack appears to be marginal, we do not think this is just statistical margin of error. Rather, we think the performance drop is meaningful in part because it is associated with a much weaker attack. More specifically, we believe that the higher severity in the acute attack is due to the different treatment of the false labels in the split MNIST experiments: while the false labels in the permuted MNIST experiments were universally set to class 0, the false labels in the split MNIST experiments are distributed over several class labels and are set to ``the first class in the two-class task''. Because the backdoor tag is intended to become associated with the false label(s), it follows that this association will be stronger when the associated is with only a small number of such false labels. In contrast, when the backdoor is associated with many labels, as in the case of the sustained attack, the association with \textit{any} of these labels becomes weaker, and the model is forced to rely on other, genuine features to make its classification decisions. At test time, the distributed nature of the false labels results in the model being less ``distracted'' by the presence of the backdoor, because the backdoor is a less-informative feature than the actual handwritten digit. Nevertheless, all three attacks demonstrate a powerful ability to hinder the model's performance, even on a task learned previously learned with near-perfect accuracy. Indeed, the ability of the attacker to have such catastrophic impact, even with a minimal amount of leverage against the model---at a point even far in the future relative to the time when the task was initially learned---should be highly worrying, as it demonstrates the lack of robustness of CL algorithms against adversarial attacks.

\section{Conclusions \& Future Work}
We have shown that elastic weight consolidation (EWC), a commonly-used regularization-based continual learning approach, is vulnerable to backdoor poisoning attacks. This vulnerability exists due to the way that EWC functions in retaining information from task-to-task. While EWC exhibits remarkable results for its intended function, an attacker may exploit the algorithm by teaching it to retain misinformation. 
We demonstrated this capability of the attacker through the utilization of \textit{backdoor} attacks, designed to create a weakness in the continual learner by which an attacker can induce a pinpoint degradation in performance at its leisure. While backdoor attacks have been previously shown to be effective at highlighting the vulnerability of stationary image recognition systems, we show for the first time that they can also be an effective tool in defeating continual learners in an online setting.

The natural immediate next steps are to explore the vulnerabilities of not only other regularization-based continual learning approaches, but also to investigate whether similar weaknesses exist in both data-based and architecture-based methods. In this effort, we only looked at a very small amount of poisoned data introduced using a backdoor pattern of fixed shape and intensity. Our future work will also include determining the impact of the amount of poisoned data in the training dataset, as well as the impact of the shape and intensity of the backdoor pattern. Perhaps equally important to demonstrating these vulnerabilities is to develop appropriate defensive solutions agains such attacks. 




An important takeaway from this work is to reemphasize the critical need for any future algorithm seeking artificial general intelligence to be cognizant to adversarial threats; this work shows that there is still much to address in the area of adversarial machine learning.
Therefore, it will be important to prioritize the development of robust learning systems, as well as defenses against adversarial attacks, such that a continual learner will be able to learn from streaming data while remaining secure against deliberate misdirection.

\section*{Acknowledgment}

This material is based upon work supported by the National
Science Foundation under grants no. 1310496 and 1429467.



\bibliographystyle{IEEEtran}
%

\bibliography{bibliography}{}

%
%

\end{document}